\documentclass[10pt,twocolumn,letterpaper]{article}

\usepackage[accsupp]{axessibility}
 \usepackage{cvpr}              % To produce the CAMERA-READY version
\usepackage{booktabs,tabularx,makecell,array}
\usepackage{booktabs}
\usepackage[table]{xcolor}
\usepackage{pifont}
\usepackage{tikz}
\usepackage{comment}
\usepackage{amsmath,amssymb}
\newcommand{\cmark}{\ding{51}} % ✓
\newcommand{\xmark}{\ding{55}} % ✗
\definecolor{cvprblue}{rgb}{0.21,0.49,0.74}
\usepackage[pagebackref,breaklinks,colorlinks,allcolors=cvprblue]{hyperref}

\def\paperID{18664} % *** Enter the Paper ID here
\def\confName{CVPR}
\def\confYear{2026}

\title{Dual-Modality Anchor-Guided Filtering for Test-time Prompt Tuning}

\author{
    Jungwon Choi \quad Eunwoo Kim\thanks{Corresponding author.} \\
    School of Computer Science and Engineering, Chung-Ang University \\
    {\tt\small \{severe0713, eunwoo\}@cau.ac.kr}
}

\begin{document}
\maketitle

\begin{abstract}

    Test-Time Prompt Tuning (TPT) adapts vision-language models using augmented views, but its effectiveness is hindered by the challenge of determining which views are beneficial. Standard entropy-based filtering relies on the internal confidence scores of the model, which are often miscalibrated under distribution shift, assigning high confidence to irrelevant crops or background regions while ignoring semantic content. To address this, we propose a dual-modality anchor-guided framework that grounds view selection in semantic evidence. We introduce a text anchor from attribute-rich descriptions, to provide fine-grained class semantics, and an adaptive image anchor that captures evolving test-time statistics. Using these anchors, we filter views based on alignment and confidence, ensuring that only informative views guide adaptation. Moreover, we treat the anchors as auxiliary predictive heads and combine their predictions with the original output in a confidence-weighted ensemble, yielding a stable supervision signal for prompt updates. Extensive experiments on 15 benchmark datasets demonstrate new state-of-the-art performance, highlighting the contribution of anchor-guided supervision as a foundation for robust prompt updates. \noindent
\textbf{Code:} \href{https://github.com/bongbong0713/DMAG-TPT}{github.com/bongbong0713/DMAG-TPT}

\end{abstract}    
\section{Introduction}
\label{sec:intro}

Vision-language models (VLMs)~\cite{CLIP, siglip, evaclip, ALIGN} have achieved remarkable zero-shot performance by aligning images and text within a shared semantic space. It enabled a wide range of visual tasks guided by natural language prompts.  Since the effectiveness of these models is highly dependent on the quality of the prompts~\cite{CLIP, Coop, du2024ipo, qu2025proapo}, recent studies have explored both hand-crafted textual templates~\cite{CLIP, ALIGN, allingham2023simple} and learnable prompts~\cite{Coop, Maple, Cocoop, CoPrompt} to improve performance. However, both approaches tend to overfit the training data distribution~\cite{ma2023understanding, Any-shift}, making it difficult to generalize to unseen domains. Test-time prompt tuning (TPT)~\cite{shu2022test} addresses this limitation by updating prompts directly on unlabeled data during inference, allowing models to better generalize to unseen domains without requiring additional labeled samples.
This setting is particularly practical, as labeled target samples are rarely available in real-world scenarios. 
%Standard TPT operates under a self-supervision signal during test time. %
For each test sample, TPT generates predictions from multiple augmented views and updates the learnable prompt by minimizing entropy of average prediction, thereby adapting the model to the current test sample.
\begin{figure}[t] % 위치 옵션: t=top, b=bottom, h=here, H=exactly here
    \centering
    \includegraphics[width=\columnwidth,
]{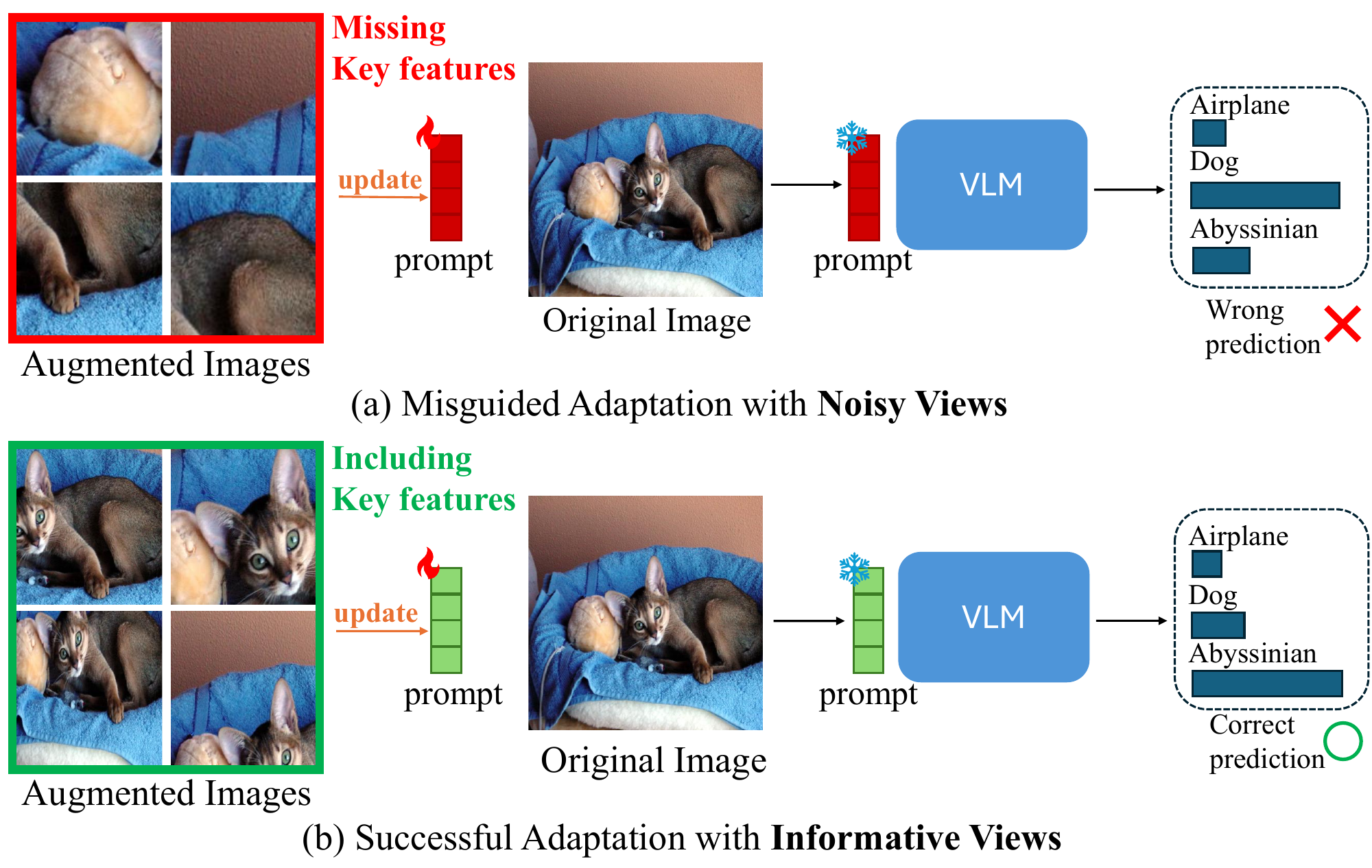}
    \vspace{-15pt}
    \caption{An example of how prompt quality varies depending on the selected views.
    When the (a) selected views are noisy and fail to capture key object features, the prompt is adapted in an incorrect direction, resulting in erroneous predictions.
    In contrast, (b) informative views containing essential features guide the prompt toward the correct semantics, resulting in accurate predictions.
    This highlights the critical role of robust view selection. }
    \vspace{-5pt}
    \label{fig:teaser}
\end{figure}
%figure pdf로

%While creating diverse views through standard augmentations is essential for the generalization of TPT methods, it can inadvertently produce noisy or uninformative views, such as background-only crops. These views risk misleading the model and degrading its performance.
However, TPT methods are vulnerable to noisy or uninformative views generated by standard augmentations (e.g., random crop or resize). Many of these views miss key object regions, over-represent the background, or include content that weakly relates to the target class. Minimizing the entropy of such unreliable views is dangerous; it reinforces confidence of the model in wrong predictions and misguides the prompt adaptation toward erroneous semantics, as shown in Fig.~\ref{fig:teaser}.
In contrast, well-selected and semantically aligned views play a crucial role; they correct the mistaken predictions of the original image and guide the prompt updates toward the true class direction. This observation underscores that the effectiveness of TPT fundamentally depends on the quality and semantic fidelity of the selected views. Despite its importance, the problem of view selection has not been largely explored in prior TPT literature. Prior methods~\cite{shu2022test, Dynaprompt, PromptAlign} rely solely on filtering views based on prediction entropy, selecting only those views that exhibit low prediction entropy. But these approaches are not sufficient, as they have two critical limitations. First, under distribution shift, prediction entropy of the model is often unreliable, leading to overconfident and inaccurate predictions~\cite{wang2024understanding, lee2024entropy, DiffTPT}. This causes semantically uninformative views to be erroneously selected. Second, entropy is typically calculated using a coarse initial prompt (e.g.,``a photo of a [CLASS]"), which lacks the ability to distinguish fine-grained attributes \cite{li2025clip, yang2025exploring} necessary to verify its identity within a confusing crop. 

One recent approach~\cite{DiffTPT} extends entropy-based filtering by measuring both entropy and the cosine similarity to the original image. Although this proves effective in that setting, the improvement largely stems from the use of diffusion~\cite{stablediff} based augmentations, which generate highly diverse views. In contrast, standard augmentations remain the practical choice for TPT methods~\cite{shu2022test, Dynaprompt, PromptAlign, C-TPT} due to their efficiency and broad applicability. However, they do not support this similarity-based criterion effectively. For example, similarity-based selection with standard augmentations reduces the diversity of views and weakens generalization, since it favors views visually similar to the original image.
Notably, we find that once view selection is performed in a semantically-aware manner, even standard augmentations can surpass diffusion-based augmentations, while avoiding the substantial computational cost of diffusion (see Section~\ref{sec:main_results}). 

%These limitations highlight a critical gap. A robust view selection mechanism must move beyond unreliable prediction entropy  and inflexible visual similarity metrics (such as the cosine similarity to the original image). It requires a filter grounded in stable, fine-grained semantic evidence, one that can reliably verify the content of diverse views without compromising the diversity of augmentations. This necessitates (1) a stable reference point (an 'anchor') that is more dependable than the unstable model predictions, and (2) a rich source of information (a 'dual-modality' approach) to capture the semantic and visual nuances that coarse prompts miss.
To address the limitations of unreliable entropy and rigid visual similarity metrics, we propose a dual-modality anchor-guided framework for robust view selection. 
%Our approach is grounded in fine-grained semantic evidence, leveraging a stable anchor as a reliable reference. %, thereby ensuring the reliable validation of diverse augmentations. 
To form a fine-grained anchor and move beyond the limitations of coarse initial prompts, we leverage large language models (LLMs) to generate attribute-rich textual descriptions that capture detailed visual properties.
Each description is weighted by its similarity to the image features and aggregated into a text anchor to better capture the semantic content of the image.
We first use the text anchors to guide initial view filtering by measuring alignment and entropy. To further incorporate domain-specific visual cues and capture appearance variations beyond textual semantics, we introduce an image anchor that is updated via the cumulative average of previously selected views. We then employ the image anchor to guide the view selection based on alignment and entropy. By combining text and image anchors, our framework captures semantic fidelity and visual consistency, enabling reliable view validation even under distribution shift where entropy and direct similarity measures often fail. This approach is also lightweight and modular, enabling seamless integration into existing prompt tuning pipelines~\cite{Coop, Cocoop, Maple} with minimal additional cost.

Finally, we leverage the filtered set of views for the prompt adaptation. However, simply minimizing the entropy of these views, a common practice in prior works~\cite{shu2022test, Dynaprompt, DiffTPT}, can encourage biased updates. To address this, we further leverage our dual anchors not only as filters but also as auxiliary predictive sources. We present a confidence-aware ensemble that dynamically weights and integrates predictions from each source based on its maximum prediction probability. This forms a highly reliable target distribution, built from the robust consensus of the dual anchors and the original prompt. The prompts are then updated by minimizing a KL-divergence loss against this target distribution, providing stable and semantically grounded supervision for robust adaptation. The main contributions are:
\begin{itemize}
    \item Our framework mitigates the limitations of entropy-based filtering by grounding the selection process in dual-modality anchors and jointly evaluating both alignment and confidence.
    
    \item We replace coarse prompts with dual-modality anchors, where the text anchor represents semantic textual information and the image anchor captures test distribution visual features.
    
    \item We leverage the anchors as predictive sources in a confidence-weighted ensemble, yielding stable supervision from augmented views.

    \item Extensive experiments across 15 diverse benchmarks demonstrate the effectiveness of our approach, achieving an average accuracy improvement of 3.36\% over all compared methods.
\end{itemize}

\section{Related Work}
\label{sec:formatting}

\subsection{Prompt Learning}
\label{subsec:prompt_learning}

Prompt learning for vision-language models (VLMs) \cite{CLIP, ALIGN} has been widely explored to enhance their performance on various downstream tasks. Inspired by prompt tuning in NLP \cite{shin2020autoprompt}, a common approach in vision tasks involves learning prompts using a small set of labeled examples \cite{Coop, Cocoop, Maple}.
CoOp \cite{Coop} adapts VLMs by optimizing a set of continuous prompt vectors in its language branch. CoCoOp \cite{Cocoop} extends CoOp \cite{Coop} by conditioning prompts on individual image features. MaPLe \cite{Maple} leverages multi-modal prompt learning with vision–language prompts. CoPrompt~\cite{CoPrompt} enforces a consistency constraint to prevent overfitting on downstream tasks. Any-Shift prompting \cite{Any-shift} introduces adaptive prompts designed to handle arbitrary distribution shifts. 

However, these prior works often rely on labeled samples from the target domain, which restricts their applicability in practical scenarios. In contrast, we tackle a more realistic test-time prompt tuning setting, where the model must adapt on-the-fly using only unlabeled test data without access to target domain labels. This paradigm is particularly important for real-world deployment, enabling robust adaptation to novel and unseen environments where labeled data is unavailable.

\subsection{Test-time Prompt Tuning}
\label{subsec:test_time_prompt_tuning}

Test-time prompt tuning (TPT) \cite{shu2022test} is an emerging paradigm that adapts vision-language models by optimizing prompts using unlabeled test samples during inference.  TPT optimizes prompts by minimizing the entropy of the averaged predictions from augmented images.  Building upon this, DiffTPT \cite{DiffTPT} incorporates a pre-trained diffusion model \cite{stablediff} to generate diverse augmentations. PromptAlign \cite{PromptAlign}, building on MaPLe \cite{Maple}, explicitly aligns the mean and variance of image token embeddings from a proxy source dataset. C-TPT \cite{C-TPT} improves test-time prompt tuning by maximizing text feature dispersion. DynaPrompt \cite{Dynaprompt} introduces a prompt queue and a scoring mechanism that enables multiple prompt updates. 

The aforementioned approaches suffer from key drawbacks: DiffTPT \cite{DiffTPT} is computationally heavy, PromptAlign \cite{PromptAlign} has limited applicability, and DynaPrompt \cite{Dynaprompt} incurs extra memory and computation overhead. More critically, they all rely on entropy-based filtering, which is semantically blind and reinforces model biases \cite{DiffTPT, AWT}.
In contrast, our dual-modality, anchor-guided filtering approach resolves these issues by creating a robust, semantically grounded supervision signal, while serving as a lightweight and versatile module that can be easily integrated into various prompt tuning frameworks.

\section{Method}
\begin{figure*}[t] % 위치 옵션: t=top, b=bottom, h=here, H=exactly here
    \centering
    \includegraphics[width=\linewidth,
    trim=5pt 80pt 0pt 0pt, 
    clip=true]{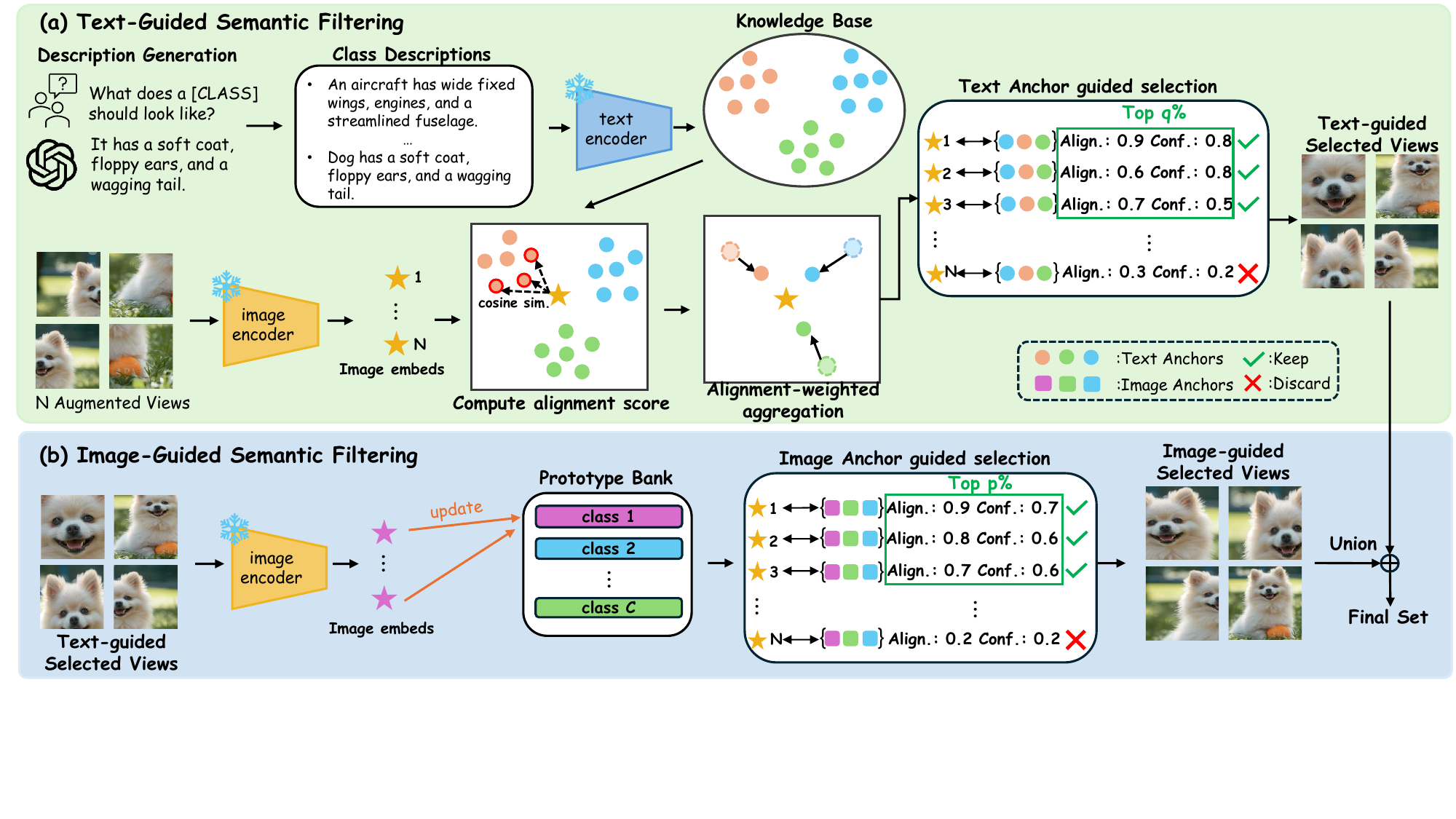}
    \vspace{-17pt}
    \caption{Overview of the proposed dual-modality anchor framework.
The process begins with (a) text anchor construction. Multiple LLM-generated descriptions per class are encoded with the CLIP text encoder, and N augmented images are encoded with the CLIP image encoder. We then compute image–text alignment by measuring cosine similarity and perform alignment-weighted aggregation to form text anchors. Using these anchors, we compute joint alignment–confidence scores to each views and keep the top q\% of N augmented views. (b) In the image anchor construction stage, the text-guided selected views are aggregated into a prototype bank that captures the visual distribution. The image anchor set is obtained by collecting the top-K class prototypes. We again compute joint alignment–confidence scores with the image anchors and retain the top p\% of N augmented views. The final view set is the union of the two filters.}
    \label{fig:method}
    \vspace{-6pt}
\end{figure*}

Standard entropy-based confidence selection \cite{shu2022test, Dynaprompt, PromptAlign, C-TPT}, retains low-uncertainty views but often fails to exclude noisy or semantically irrelevant augmentations. To address this, we introduce a dual-modality anchor mechanism that guides view selection with semantic evidence as illustrated in Fig.~\ref{fig:method}. Our approach leverages text anchors derived from attribute-rich descriptions and adaptive image anchors accumulated from test-time embeddings, forming the basis for a more robust and informative selection process.

\subsection{Text-Guided Semantic Filtering}
\label{sec:llm_filtering}

To enhance the discriminative alignment between visual inputs and textual descriptions, we propose a filtering method that utilizes LLM-generated descriptions. We treat these descriptions as proxy attributes that are both \textit{semantically rich} and \textit{class-specific}. This approach of leveraging fine-grained semantic attributes over coarse, label-only prompts can be aligned with recent work in attribute-based recognition~\cite{pratt2023does, maniparambil2023enhancing, naeem2023i2mvformer, qu2025proapo}, which has also highlighted their importance in enhancing discriminative power.

\paragraph{Attribute-Rich Description Generation.}
\vspace{-18pt}
To enrich the textual representation, we leverage LLMs to generate attribute-oriented descriptions for each class. 
Following the practice in \cite{pratt2023does}, we prompt an LLM with a diverse set of instructions, such as
\textit{``Describe what a(n) [CLASS] looks like''}.
This process yields multiple attribute-rich descriptions per class, which are encoded by the text encoder into a knowledge base of textual attributes.
This knowledge base provides stronger semantic grounding than conventional template prompts \cite{CLIP, ALIGN} and serves as the foundation for our filtering process. 
All descriptions are generated and encoded offline as a one-time preprocessing step; during test time, our method only relies on lightweight similarity computations with pre-encoded text features, incurring negligible additional cost.
\vspace{-15pt}

\paragraph{Alignment-Weighted Aggregation.}
LLM-generated descriptions provide diverse semantically rich attributes, but their relevance varies depending on the specific visual instance. To effectively prioritize descriptions aligned with the actual image content, we employ an adaptive aggregation mechanism. We are given image features from the augmented views, denoted as $\{\mathbf{e}_b\}_{b=1}^{B}$, where $B$ is the number of views. 
Similarly, we have text description features $\{\mathbf{w}_c^i\}_{i=1}^{N}$, where $N$ is the number of descriptions for each class $c$. All features are obtained from CLIP encoders ~\cite{CLIP}. For each tuple $\rho \triangleq (b, c, i)$, we compute an alignment score: $s_\rho = \operatorname{sim}(\mathbf{e}_b, \mathbf{w}^i_c) - \operatorname{sim}(\mathbf{e}_b, \bar{\mathbf{w}}_c)$,
where $\operatorname{sim}(\cdot, \cdot)$ denotes cosine similarity and 
$\bar{\mathbf{w}}_c = \frac{1}{N} \sum_{i=1}^{N} \mathbf{w}^i_c$ is the class-wise mean feature. This score quantifies the \textit{marginal benefit} of a specific description $\mathbf{w}_c^i$ over the generic, class-wise mean feature $\bar{\mathbf{w}}_c$. A high score indicates that the description provides semantic details that are particularly relevant to the image view. 

To ensure a discriminative signal even under low-variance scenarios, we convert these scores into per-view weights $a_{b,c,i}$ via a softmax operation: $a_{b,c,i} = \frac{\exp(s_{b,c,i})}{\sum_{j=1}^{N} \exp(s_{b,c,j})}$.
These weights represent the relevance of description $i$ (for class $c$) to a specific view $b$. 
We then average these weights across all $B$ views to obtain the final class-level description weights $\bar{a}_{c,i} = \frac{1}{B} \sum_{b=1}^{B} a_{b,c,i}$.
Finally, the text anchor $\mathbf{t}_c$ is formed by an alignment-weighted aggregation:
$
    \mathbf{t}_c = \sum_{i=1}^{N} \bar{a}_{c,i} \mathbf{w}_c^i.
$
Note that this entire process is performed adaptively for every single test image. \
\vspace{-10pt}

\paragraph{Joint Alignment-Confidence Scoring.}
We define a composite score for each view by augmenting an entropy-based uncertainty measure with a semantic alignment score. For a given view feature $\mathbf{e}$ and the number of class $C$, we compute two components: alignment and confidence.
\begin{equation}
s_{\text{align}}(\mathbf{e}) = \max_{c \in \mathcal{C}} 
    \operatorname{sim}\!\left(\mathbf{e}, \mathbf{t}_c\right), \quad
s_{\text{conf}}(\mathbf{e}) = 1 - 
    \frac{\operatorname{H}\!\left(p(y \mid \mathbf{e})\right)}{\log C}.
\label{eq:scores}
\end{equation}
$\operatorname{H}(\cdot)$ denotes the prediction entropy, and the normalized entropy $\frac{\operatorname{H}(p(y \mid \mathbf{e}))}{\log C} \in [0,1]$ measures the uncertainty of the per-view predictive distribution:
$p(y \mid \mathbf{e}) = \operatorname{Softmax}\!\left(\tau \cdot \mathbf{e}^{\top} \mathbf{t}_c\right)$
where $y$ is class label, $\tau$ is a learned temperature parameter, and the probabilities are computed from the similarity between the image feature and text anchors. The final score is a weighted sum of alignment and confidence:
\begin{equation}
    \operatorname{S}_{\text{text}}(\mathbf{e}) = \alpha_1 s_{\text{align}}(\mathbf{e}) + \alpha_2 s_{\text{conf}}(\mathbf{e}),
\label{eq:textscore}
\end{equation}
where the coefficients $(\alpha_1, \alpha_2)$ balance the trade-off between semantic alignment and uncertainty.
We retain the top-$q\%$ of views with the highest $\operatorname{S}{_\text{text}}$.
This replaces the entropy-based rule with a criterion that requires both strong alignment to some class anchors and low predictive uncertainty, thereby down-weighting confidently wrong or background-dominated views.

\subsection{Image-Guided Semantic Filtering}
\label{sec:img_filtering}

Our text anchor effectively adapts a semantic prior to the specific view. To complement this semantic-level adaptation, we introduce an image anchor strategy designed to capture the overall visual statistics of the test-time distribution. Specifically, we maintain a class-wise prototype bank that continuously aggregates representative visual features during test-time inference, serving as a memory of how each class appears under the current distribution. From this bank, the image anchor is constructed to provide a compact and dynamically updated representation of the visual domain.
\vspace{-10pt}

\paragraph{Image Anchor Construction.}
At test time, views selected by the text-guided filter are used to incrementally update a class-wise prototype bank $\mathcal{P} = \{\mathbf{p}_c\}_{c=1}^C$. For each selected embedding $\mathbf{e}$, we obtain its predicted class label $c$ from the original model before adaptation. The corresponding prototype $\mathbf{p}_{c}$ is updated via cumulative averaging, $\mathbf{p}_{c} \leftarrow \frac{n_c \mathbf{p}_c + \mathbf{e}}{n_c + 1}$,
where $n_c$ is the number of past updates for class $c$. For a given test image, we compute class scores $\bar{\pi}_c$ by averaging predictions over selected views and choose the top-$K$ class indices, $\mathcal{C}_K = \operatorname{TopK}(\{\bar{\pi}_c\})$. The image anchor set $\mathcal{A}_{\text{img}}$ is then formed by gathering the corresponding prototypes:
$
    \mathcal{A}_{\text{img}} = \{ \mathbf{p}_c \mid c \in \mathcal{C}_K \} .
$
\vspace{-5pt}

\paragraph{Joint Alignment-Confidence Scoring.}
In parallel to the text-guided filtering, we introduce a second stage of view selection driven by image anchors. For all aug views, alignment with the image anchor set and predictive confidence are computed similar to Eq.~\ref{eq:scores}.
These terms are combined into a single score:
\begin{equation}
    \operatorname{S}_{\text{img}}(\mathbf{e}) = \beta_1 s_{\text{align}}(\mathbf{e}) + \beta_2 s_{\text{conf}}(\mathbf{e}) ,
\label{eq:imagescore}
\end{equation}
with $(\beta_1, \beta_2)$ balancing alignment and confidence.
We retain the top-$p\%$ of views with the highest $\operatorname{S}{_\text{img}}$. The final set of selected views is the union of both filters, with any duplicates counted once.

\subsection{Adaptive Ensemble of Multimodal Predictions}
\label{sec:adaptive_ensemble}

To further enhance stability and accuracy for these selected views, we present a confidence-aware weighting ensemble that dynamically integrates predictions from multiple sources. This approach mitigates the influence of noisy predictions by assigning higher weights to more confident ones, thus generating a refined training signal. The model produces three distinct similarity vectors for each selected view, obtained by computing the cosine similarity between the image feature and (i) the original prompt embeddings, (ii) the text anchors $\mathbf{t}_c$, and (iii) the image prototypes $\mathbf{p}_c$. We denote these as $\mathbf{z}_{\text{prompt}}$, $\mathbf{z}_{\text{text}}$, and $\mathbf{z}_{\text{image}}$, respectively, and refer to the set of sources as $S = \{\text{prompt, text, image}\}$.

To integrate complementary evidence from these sources, we aggregate their outputs based on their predictive confidence. First, we compute the probability distribution for each source as 
$\mathbf{q}_k = \operatorname{Softmax}(\mathbf{z}_k), \; k \in S$. Next, we define the confidence score $\gamma_k$ for each source as the maximum probability in its distribution: $\gamma_k = \max(\mathbf{q}_k)$. These confidence scores are then normalized to produce the ensemble weights. The weight $w_k$ for each source $k \in S$ is:
$
    w_k = \frac{\gamma_k}{\sum_{j \in S} \gamma_j + \epsilon},
$
where $\epsilon$ is a small constant for numerical stability. Finally, the ensemble vector is generated by a weighted sum of the individual scores:
\begin{equation}
    \mathbf{z}_{\text{ens}} = \sum_{k \in S} w_k \mathbf{z}_k.
\end{equation}
This confidence-aware ensemble creates a de-biased supervision signal by forming a robust consensus that leverages the most reliable predictive head for each image, thereby mitigating single-source errors.
\vspace{-10pt}

\paragraph{Target Distribution Sharpening.}

We compute per-view probabilities via softmax of the ensemble logits ${z}_{\text{ens}}$, average it over the selected views, and sharpen it using temperature scaling ($T=0.3$) to form the target distribution, $\tilde{\mathbf{q}}$. 
The original prediction, $\mathbf{p}_{v}$, is formed by averaging the original prompt logits over the same views and applying a softmax function. We then update the prompts by minimizing the KL divergence:
\begin{equation}
    \mathcal{L}_{\text{TPT}} = D_{\mathrm{KL}}(\tilde{\mathbf{q}} \,\|\, \mathbf{p}_{v}).
\end{equation}
This loss encourages the prompt-driven prediction $\mathbf{p}_{v}$ to match the confident, ensemble-based target distribution $\tilde{\mathbf{q}}$. Note that prior TPT methods~\cite{shu2022test, DiffTPT, Dynaprompt} directly minimize the entropy of the averaged distribution, pushing the model toward self-sharpening. Our method stabilizes adaptation by deriving a sharpened target distribution from the ensemble, which prevents collapse into overconfident yet semantically inconsistent predictions.
\section{Experiments}

\begin{table*}[t]
\caption{Comparison of methods on multiple ImageNet variants. We report top-1 accuracy (\%) on each dataset and their averages. Numbers in parentheses indicate improvement over the corresponding baseline
(TPT, CoOp+TPT, or MaPLe+TPT).}
\vspace{-5pt}
\centering
\resizebox{\textwidth}{!}{%
\begin{tabular}{l c c c c c c c c}
\hline
\textbf{Method} & \textbf{ImageNet} & \textbf{ImageNet-A} & \textbf{ImageNet-V2} & \textbf{ImageNet-R} & \textbf{ImageNet-Sketch} & \textbf{Average} & \textbf{OOD Avg.} \\
\hline
CLIP-ViT-B/16 & 66.73 & 47.87 & 60.86 & 73.98 & 46.09 & 59.11 & 57.20 \\
\hline
\multicolumn{8}{c}{\textit{Prompt learning methods without test-time tuning}} \\
\hline
CoOp & 71.51 & 49.71 & 64.20 & 75.21 & 47.99 & 61.72 & 59.28 \\
CoCoOp & 71.02 & 50.63 & 64.47 & 76.18 & 48.75 & 62.13 & 59.91 \\
MaPLe & 70.72 & 49.15 & 64.07 & 76.98 & 50.90 & 62.36 & 60.28 \\
CoPrompt & 70.80 & 49.43 & 64.25 & 77.51 & 50.50 & 62.50 & 60.42 \\
Any-shift Prompt & -- & 49.80 & 64.53 & 77.56 & 51.52 & -- & 60.85 \\
\hline
\multicolumn{8}{c}{\textit{Test-time prompt tuning methods}} \\
\hline
TPT & 68.98 & 54.77 & 63.45 & 77.06 & 47.06 & 62.06 & 60.81 \\
DiffTPT & 70.30 & 55.68 & 65.10 & 75.00 & 46.80 & 62.26 (+0.20) & 60.65 (-0.16) \\
C-TPT & 69.30 & 52.90 & 63.40 & 78.00 & 48.50 & 62.42 (+0.36) & 60.70 (-0.11) \\
DynaPrompt & 69.61 & 56.17 & 64.67 & 78.17 & 48.22 & 63.37 (+1.31) & 61.81 (+1.00) \\
\rowcolor{gray!15}
\textbf{Ours} & \textbf{72.21} & \textbf{59.65} & \textbf{65.35} & \textbf{80.25} & \textbf{51.24} & \textbf{65.74 (+3.68)} & \textbf{64.12 (+3.31)} \\
\hline
CoOp + TPT & 73.61 & 57.95 & 66.83 & 77.27 & 49.29 & 64.99 & 62.84 \\
CoOp + DynaPrompt & \textbf{74.08} & 60.55 & 67.25 & 79.15 & 50.28 & 66.26 (+1.27) & 64.31 (+1.47) \\
\rowcolor{gray!15}
\textbf{CoOp + Ours} & 74.03 & \textbf{62.40} & \textbf{67.50} & \textbf{81.02} & \textbf{52.58} & \textbf{67.51 (+2.52)} & \textbf{65.88 (+3.04)} \\
\hline
MaPLe + TPT & 71.87 & 58.08 & 64.87 & 78.12 & 48.16 & 64.22 & 62.31 \\
MaPLe + PromptAlign & - & 59.37 & 65.29 & 79.33 & 50.23 & - & 63.56 (+1.25) \\
MaPLe + DynaPrompt & \textbf{72.71} & \textbf{60.72} & 66.34 & 79.57 & 50.25 & 65.92 (+1.70) & 64.22 (+1.91) \\
\rowcolor{gray!15}
\textbf{MaPLe + Ours} & 72.31 & 59.65 & \textbf{66.84} & \textbf{80.50} & \textbf{52.11} & \textbf{66.28 (+2.06)} & \textbf{64.78 (+2.47)} \\
\hline
\end{tabular}
}
\vspace{-3pt}
\label{tab:tpt_aug_comparison}
\end{table*}

\subsection{Experimental Setup}

\textbf{Datasets.} Following previous methods \cite{shu2022test, DiffTPT, Dynaprompt}, we evaluate our method in two settings: domain generalization and cross-dataset transfer. For domain generalization, we use ImageNet \cite{imagenet} along with four distribution-shifted test sets: ImageNet-A \cite{imagenet-a},  ImageNet-R \cite{imagenet-r}, ImageNet-V2 \cite{imagenet-v2},  and ImageNet-Sketch \cite{imagenet-sketch}. For cross-dataset evaluation, we consider 10 diverse image classification benchmarks: Flower102 \cite{Flower102}, DTD \cite{DTD}, OxfordPets \cite{OxfordPets}, StanfordCars \cite{Cars}, UCF101 \cite{UCF101}, Caltech101 \cite{Caltech101}, Food101 \cite{Food101}, SUN397 \cite{SUN397}, FGVC-Aircraft \cite{Aircraft}, and EuroSAT \cite{eurosat}. These datasets cover a diverse spectrum of visual domains, ranging from fine-grained object classification to satellite imagery.

\noindent \textbf{Implementation details.} Our model is built upon CLIP \cite{CLIP} with a ViT-B/16 backbone. We optimize four learnable prompt tokens in text embedding space.  Following TPT \cite{shu2022test}, we generate 63 random resized crops per test image and include the original image, yielding a batch of 64 views. For each view, we compute entropy and alignment with both text and image anchors, retaining the top 10\% for text anchors (6 views) and the top 5\% for image anchors (3 views). The final set of selected views is the union of these two sets, resulting in a total of 6 to 9 unique views depending on the degree of overlap. Prompt optimization is performed using KL divergence against the sharpened predictions of the selected augmented views. We use AdamW with a learning rate of 0.003. All experiments were conducted using a single NVIDIA RTX 6000 ada GPU. All prompt tuning methods are trained on ImageNet with 16-shot.

\subsection{Main Results}
\label{sec:main_results}

\textbf{Comparisons on domain generalization.} We evaluate our approach under the domain generalization setting by comparing it against both prompt learning and test-time prompt tuning baselines, as summarized in Table~\ref{tab:tpt_aug_comparison}. The metric \textit{Average} is computed as the mean top-1 accuracy over all five datasets.
The metric \textit{OOD Average} is computed as the mean top-1 accuracy over the four out-of-distribution datasets, excluding ImageNet. Prompt learning methods, such as CoOp, CoCoOp, and MaPLe, are trained with the supervised cross-entropy loss on ImageNet, while test-time prompt tuning methods, including TPT, DiffTPT, and DynaPrompt, adapt prompts at inference time using unlabeled test data. 

Our method consistently improves performance across all metrics. In particular, we achieve an average accuracy of 65.74 and an OOD average of 64.12, outperforming TPT by +3.68\% and +3.04\%, respectively. Compared to the strongest baseline, DynaPrompt, our method still provides a clear margin of +2.37\% in average and +2.31\% in OOD accuracy. Since our method can be seamlessly integrated into existing prompt tuning frameworks, combining it with CoOp and MaPLe further improves performance. Across both bases, our method delivers the best average and OOD average: with CoOp, improvements of +2.52\% in average and +3.04\% in OOD over TPT; with MaPLe, improvements of +2.06\% and +2.47\% over TPT. Notably, even without ImageNet-pretrained prompts from CoOp, Ours alone surpasses OOD accuracy of CoOp+TPT, highlighting that our anchor-guided filtering is complementary to test-time prompt tuning. 

We further analyze the inference efficiency of our method in comparison with other test-time prompt tuning approaches, as shown in Table~\ref{tab:tpt_time_comparison}. The evaluation is performed on the ImageNet-R dataset. While TPT achieves modest improvements with low overhead, its overall gain is limited. DiffTPT attains lower accuracy and already takes over 0.6 sec per image due to multiple prompt-update steps; latency increases further when diffusion is used. DynaPrompt also improves accuracy, yet incurs additional overhead (0.4 sec) due to optimizing and maintaining multiple prompts simultaneously. Our method attains the highest accuracy with negligible overhead (0.2 sec), almost same runtime as vanilla TPT. These results confirm that our anchor-guided filtering introduces minimal computational burden while delivering larger gains.

\begin{table}[t]
\centering
\caption{Comparison of test-time tuning methods. Accuracy, gain over the CLIP baseline, and test time are reported.}
\vspace{-5pt}
\begin{tabular}{lccc}
\hline
\textbf{Method} & \textbf{Accuracy} & \textbf{Gain} & \textbf{Testing Time}\\
\hline
CLIP              & 73.98 & - &\\
TPT         & 77.06 & +3.06 &0.2 sec\\
DiffTPT  & 75.00 & +1.02&$>$ 0.6 sec \\
DynaPrompt & 78.17 & +4.19&0.4 sec \\
\hline
\textbf{Ours}      & \textbf{80.25} & \textbf{+6.27} & 0.2 sec\\
\hline
\end{tabular}
\label{tab:tpt_time_comparison}
\vspace{-5pt}
\end{table}

\begin{table*}[t]
\centering
\caption{Comparison of methods on multiple downstream datasets. }
\vspace{-5pt}
\small
\resizebox{\textwidth}{!}{%
\begin{tabular}{l*{11}{c}}
\hline
\textbf{Method} & 
\textbf{Flower102} & 
\textbf{DTD} & 
\textbf{Pets} & 
\textbf{Cars} & 
\textbf{UCF101} & 
\textbf{Caltech101} & 
\textbf{Food101} & 
\textbf{SUN397} & 
\textbf{Aircraft} & 
\textbf{EuroSAT} & 
\textbf{Average} \\
\hline
CLIP-ViT-B/16  & 67.44 & 44.27 & 88.25 & 65.48 & 65.13 & 93.35 & 83.65 & 62.59 & 23.67 & 42.01 & 63.58 \\
\hline
\multicolumn{12}{c}{\textit{Prompt learning methods without test-time tuning}} \\
\hline
CoOp     & 68.71 & 41.92 & 89.14 & 64.51 & 65.55 & 93.70 & 85.30 & 58.15 & 18.47 & 46.39 & 63.88 \\
CoCoOp   & 70.85 & 45.45 & 90.46 & 64.90 & 68.44 & 93.79 & 83.97 & 66.89 & 22.29 & 39.23 & 64.63 \\
MaPLe    & 72.23 & 46.49 & 90.49 & 65.57 & 68.69 & 93.53 & 86.20 & 67.01 & 24.74 & 48.06 & 66.30 \\
CoPrompt & 72.30 & 47.07 & 90.73 & 65.67 & 69.73 & 94.50 & 86.43 & 67.57 & 24.00 & 51.90 & 67.00 \\
\hline
\multicolumn{12}{c}{\textit{Test-time prompt tuning methods}} \\
\hline
TPT      & 68.98 & 47.75 & 87.79 & 66.87 & 68.04 & 94.16 & 84.67 & 65.50 & 24.78 & 42.44 & 65.10 \\
DiffTPT     & 70.10 & 47.00 & 88.22 & 67.01 & 68.22 & 92.49 & \textbf{87.23} & 65.74 & 25.60 & 41.04 & 65.47 (+0.37) \\
C-TPT       & 69.80 & 46.00 & 88.20 & 65.80 & 65.70 & 93.60 & 83.70 & 64.80 & 24.00 & 43.20 & 64.80 (-0.30) \\
DynaPrompt  & 69.95 & 47.96 & 88.28 & 67.65 & 68.72 & 94.32 & 85.42 & 66.32 & 24.33 & 42.28 & 65.52 (+0.42) \\
\rowcolor{gray!15}
\textbf{Ours}         & \textbf{74.71} &	\textbf{53.19}&	\textbf{89.94}&	\textbf{68.75}&	\textbf{72.24}&	\textbf{94.85}&	85.45&	\textbf{70.28}&	\textbf{28.11}&	\textbf{50.62}& \textbf{68.81 (+3.71)} \\
\hline
CoOp + TPT &68.48 & 43.32 & 89.48 & 66.77 & 68.91 & 93.15 & \textbf{86.48} & 66.06 & 20.51 & 37.73& 64.09
\\
CoOp + DynaPrompt & 69.38 & 46.98 & \textbf{90.04} & 67.35 & 69.54 & 94.40 & 86.45 & 66.17 & 21.35 & 38.55 & 65.02 (+0.93)
\\
\rowcolor{gray!15}
\textbf{CoOp + Ours} & \textbf{74.1}	&\textbf{49.59}&	88.85&	\textbf{68.83}&	\textbf{71.24}&	\textbf{94.60}&	84.42&	\textbf{69.73}&	\textbf{23.73}&	\textbf{50.37}& \textbf{67.55 (+3.46)}
\\
\hline
MaPLe + TPT    & 72.37 & 45.87 & 90.72 & 66.50 & 69.19 &  93.59 &  86.64 &  67.54 & 24.70 & 47.80 & 66.50 
\\
MaPLe + PromptAlign    & 72.39 & 47.24 & 90.76 & 68.50 & 69.47 &  94.01 &  86.65 &  67.54 & 24.80 & 47.86 & 66.92 (+0.42)
\\
MaPLe + DynaPrompt    & 73.28 & 48.75 & \textbf{90.95} & 68.26 & 69.85 &  \textbf{95.17} &  86.60 &  68.18 & 24.36 & 47.53 & 67.29 (+0.79)
\\
\rowcolor{gray!15}
\textbf{MaPLe + Ours}    & \textbf{75.10} & \textbf{53.72} & 90.32 & \textbf{69.55} & \textbf{72.21} & 94.16 & \textbf{87.41} & \textbf{70.32} & \textbf{29.11} & \textbf{51.69} & \textbf{69.36 (+2.86)} \\
\hline
\end{tabular}
}
\label{tab:downstream_results}
\vspace{-5pt}
\end{table*}

\begin{table*}[t]
\centering
\caption{Comparison of selection strategies.
We compare the conventional confidence-based selection with the proposed anchor-guided selection. 
The last row reports the performance of our method.}
\vspace{-5pt}
\resizebox{\linewidth}{!}{%
\begin{tabular}{l c c c c c c c c}
\toprule
\textbf{Method} & \textbf{ImageNet} & \textbf{ImageNet-A} & \textbf{ImageNet-V2} & \textbf{ImageNet-R} & \textbf{ImageNet-Sketch} & \textbf{Average} & \textbf{OOD Avg.} \\
\midrule
CLIP (ViT-B/16)   & 66.73 & 47.87 & 60.86 & 73.98 & 46.09 &59.11& 57.20 \\
\midrule
TPT (baseline)    & 68.47 & 51.12 & 62.24 & 75.54 & 47.20 & 60.91&59.03 \\
+ Cosine Sel. (Orig. Img.) & 69.02 & 53.67  & 63.37  & 77.00  & 48.00  & 62.21 (+1.30) & 60.51 (+1.48) \\
+ Confidence Sel. & 68.98& 54.77  & 63.45  & 77.06  & 47.94  & 62.44 (+1.53)&60.81 (+1.78) \\
\rowcolor{gray!15}
+ \textbf{Ours (Anchor Sel.)} 
                  &\textbf{70.53 }
                  &\textbf{56.83 } & \textbf{64.54 } & \textbf{79.07 } & \textbf{49.05 } & 
                  \textbf{64.00 (+3.09)}& \textbf{62.37 (+3.34)} \\
\bottomrule
\end{tabular}%
}
\label{tab:selection}
\vspace{-5pt}
\end{table*}

\noindent \textbf{Comparisons on cross-dataset.} Table \ref{tab:downstream_results} presents the top-1 accuracy comparison of our method against prompt learning and test-time prompt tuning methods in 10 downstream datasets. 
A critical observation from the table is the performance trade-off in test-time tuning. Many existing TPT methods show lower average performance than the best prompt learning methods, such as MaPLe and CoPrompt, that do not perform test-time adaptation. This highlights a significant challenge where test-time adaptation can struggle to improve performance or fail to provide substantial gains over a strong, pretrained prompt. The proposed method successfully overcomes this limitation. It achieves an average accuracy of 68.81\%, which not only surpasses all other TPT methods by a large margin but also significantly outperforms the strongest prompt learning method, CoPrompt, by 1.81\%.
Furthermore, our method demonstrates strong synergy when combined with MaPLe. The MaPLe + Ours variant achieves a new state-of-the-art average accuracy of 69.38\%. This represents a substantial 2.88\% improvement over the baseline MaPLe + TPT, proving superior effectiveness of our method as a test-time tuning strategy.

\subsection{More Studies}
\vspace{-2pt}
\noindent \textbf{Comparisons of selection strategies. } We compare several view selection approaches to examine their impact on test-time prompt tuning.
\textit{TPT (baseline)} does not apply any selection mechanism.
\textit{Cosine Sel. (Orig. Img.)} incorporates both the prediction entropy and the cosine similarity to the original image as selection criteria.
\textit{Confidence sel} only incorporates prediction entropy, selecting views that show low prediction entropy.
The cosine selection method showed the lowest performance, as shown in Table~\ref{tab:selection}, demonstrating that simple similarity to the original image alone cannot serve as a reliable semantic indicator.
Our \textit{Anchor Sel.} attains 64.00\% in average and 62.37\% in OOD average accuracy, outperforming TPT by +3.09\% and +3.34\%, respectively. This demonstrates that effective view selection is key to TPT.

% in document
\begin{table}[t]
\centering
\caption{Ablation study on four components and the results on three benchmarks.}
\vspace{-5pt}
\small
\setlength{\tabcolsep}{3.5pt}
\renewcommand{\arraystretch}{1.15}
\begin{tabular}{ccccccc}
\toprule
$\mathbf{S_{\text{text}}}$ &
$\mathbf{S_{\text{image}}}$ &
$\mathbf{z_{\text{text}}}$ &
$\mathbf{z_{\text{image}}}$ &
\textbf{Flowers} & \textbf{DTD} & \textbf{ImageNet} \\
\midrule
\xmark & \xmark & \xmark & \xmark & 69.39 & 46.63 & 68.98 \\
\cmark & \xmark & \xmark & \xmark & 70.36 & 49.41 & 70.52 \\
\xmark & \cmark & \xmark & \xmark & 70.08 & 47.75 & 70.01 \\
\cmark & \cmark & \xmark & \xmark & 71.54 & 49.70 & 70.53 \\
\cmark & \cmark & \cmark & \xmark & 72.84 & 51.36 & 71.04 \\
\cmark & \cmark & \xmark & \cmark & 72.76 & 51.47 & 71.34 \\
\rowcolor{gray!15}
\cmark & \cmark & \cmark & \cmark & \textbf{74.71} & \textbf{53.19} & \textbf{72.21} \\
\bottomrule
\end{tabular}
\vspace{-5pt}
\label{tab:ablation}
\end{table}
\newcolumntype{Y}{>{\centering\arraybackslash}X}  % centered X column

\noindent \textbf{Effectiveness of Text and Image Anchors.} To analyze the contribution of each component in our dual-modality framework, we conduct an ablation study by incrementally activating the text and image anchors for both filtering and ensemble stages, as summarized in Table \ref{tab:ablation}.
Starting from the baseline TPT~\cite{shu2022test} (\xmark\ in all rows), introducing text anchor filtering ($S_{\text{text}}$) improves accuracy across all datasets, verifying that text anchor helps retain informative views. Adding image anchor filtering ($S_{\text{image}}$) leads to additional improvements, confirming that image anchor provides complementary cues for view quality.
When we additionally incorporate text anchor predictions ($z_{\text{text}}$) and image anchor predictions ($z_{\text{image}}$) in the ensemble, performance consistently improves.
The full configuration achieves the best results, highlighting the effectiveness of the proposed dual-modality anchor-guided framework.

% 1:0 이랑 0:1 추가

\noindent \textbf{Sensitivity to hyperparameters.} We study the influence of the weighting ratios in Eqs.~\ref{eq:textscore} and ~\ref{eq:imagescore} that control the trade-off between the alignment and entropy terms in our joint scoring modules.
Results in Table~\ref{tab:ratio_ablation} show that the text-anchor module attains its best accuracy at 1:2 (alignment:entropy). The image-anchor module is optimal at 2:1.%, implying that entropy provides a more informative signal for text-conditioned scoring. Conversely, the image-anchor module is optimal at 2:1, highlighting that alignment contributes more strongly when scoring with image anchors.

\noindent \textbf{Effectiveness of alignment-weighted aggregation.} We investigate the impact of alignment-weighted aggregation, which adaptively fuses multiple LLM-generated text descriptions into a single anchor based on their semantic relevance to the image. Without aggregation, the anchor is formed by simply averaging all description embeddings. 
The simple averaged text anchor achieves 62.53\% OOD accuracy, while the alignment-weighted aggregated text anchor achieves 63.01\%. Incorporating alignment-weighted aggregation thus improves OOD accuracy by 0.48\%, confirming that weighting descriptions by their alignment scores yields more faithful and informative anchors.

\begin{table}[t]
\centering
\caption{Effect of hyperparmeters. }
\vspace{-5pt}
\small
\begin{tabular}{c|ccccc}
\toprule
$\alpha_1 : \alpha_2$ & 1:4 & 1:2 & 1:1 & 2:1 & 4:1 \\
\midrule
OOD Avg (\%) & 62.73 & \textbf{63.01} & 62.95 & 62.87 & 62.66 \\
\bottomrule
\end{tabular}

\vspace{0.8em} % ✅ 두 테이블 사이 간격

\begin{tabular}{c|ccccc}
\toprule
$\beta_1 : \beta_2$ & 1:4 & 1:2 & 1:1 & 2:1 & 4:1 \\
\midrule
OOD Avg (\%) & 62.75 & 62.83 & 62.95 & \textbf{63.01} & 62.80 \\
\bottomrule
\end{tabular}
\label{tab:ratio_ablation}
\vspace{-5pt}
\end{table}

\begin{table}[t]
\centering
\caption{
Effect of loss and ensemble strategy on cross-dataset (CD) and out-of-distribution (OOD) performance.
}
\vspace{-5pt}
\label{tab:loss_hyper}

% p{width} + \arraystretch 조합으로 overflow 방지
\renewcommand{\arraystretch}{1.05}
\setlength{\tabcolsep}{4pt}
\small
\begin{tabular}{lcc}
\toprule
\textbf{Setting} & \textbf{OOD (\%)} & \textbf{CD (\%)} \\
\midrule
EM + Simple Avg. & 61.15 & 66.72 \\
EM + Conf. Ens. & 62.23 & 67.56 \\
KLD + Simple Avg. & 62.51 & 67.53 \\
\rowcolor{gray!15}
KLD + Conf. Ens. (Ours) & \textbf{63.01} & \textbf{68.81} \\
\bottomrule
\end{tabular}
\label{tab:loss_ensemble}
\vspace{-5pt}
\end{table}

\noindent \textbf{Effect of loss design and ensemble strategy. } We analyze the contribution of our sharpened KLD loss and the confidence-aware ensemble used during test-time adaptation.
As shown in Table~\ref{tab:loss_ensemble}, replacing the standard entropy minimization (EM) loss with the KLD objective improves both cross-dataset (CD) and out-of-distribution (OOD) accuracy.
While simple averaging of predictions provides marginal gains, incorporating confidence-aware weighting yields further improvements by stabilizing the target distribution.
The combination of the KLD loss and confidence-aware ensemble achieves the best overall performance, showing that the two components are complementary.

\subsection{Qualitative Analysis of View Selection}

\begin{figure}[t]
    \centering
    \includegraphics[width=\columnwidth,
]{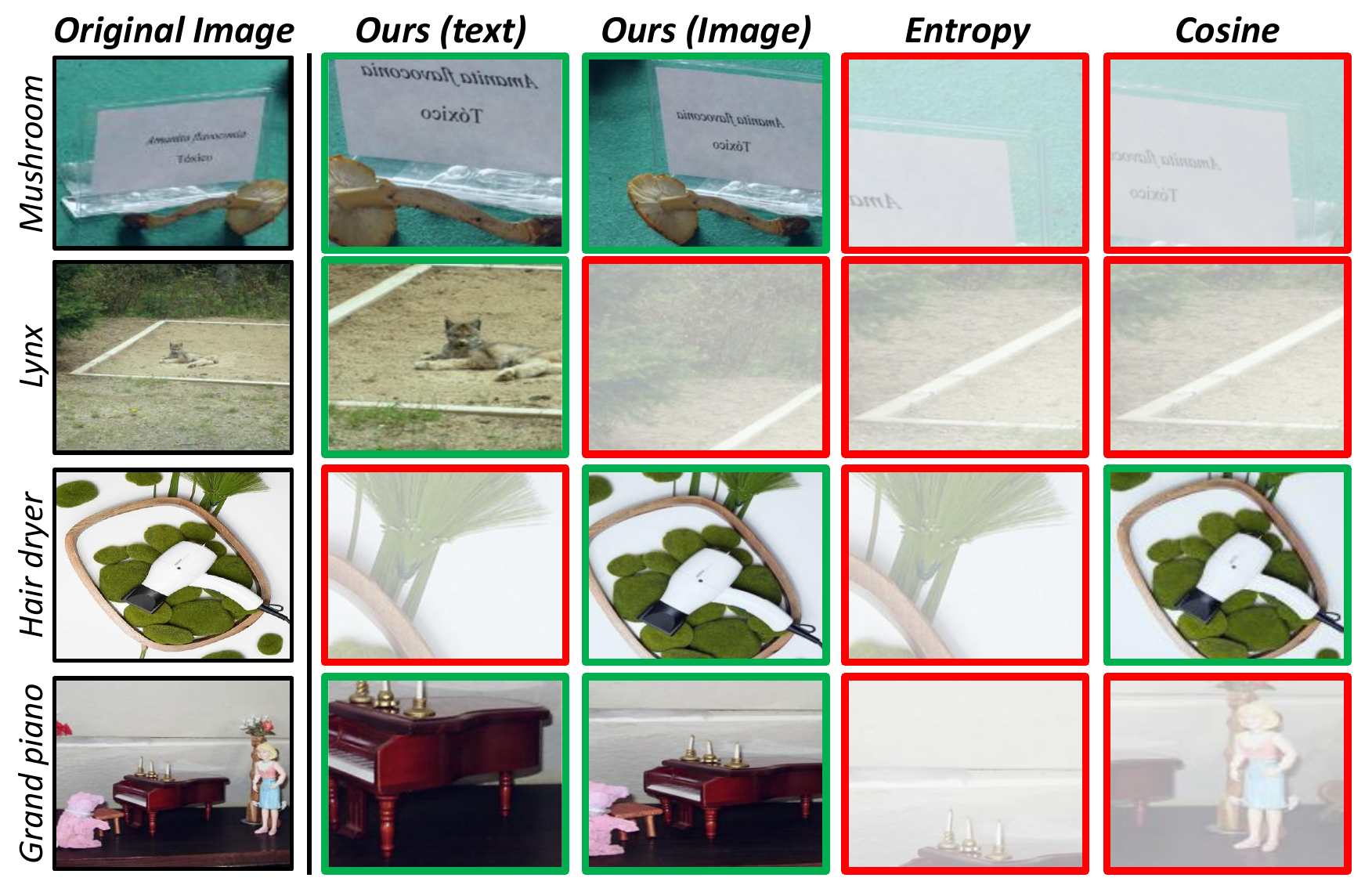}
    \vspace{-15pt}
    \caption{
Visualization of views selected by each anchor modality.
For each test image, we compare the augmented views selected by the text-guided and image-guided filters with the previous selection strategies. Green boxes indicate correctly selected views containing the target object, whereas red boxes mark views where the object is absent or unrecognizable.
    }
    \label{fig:selection}
    \vspace{-10pt}
\end{figure}

To better understand the role of each selection component, we qualitatively compare the augmented views retained by different filtering strategies in Fig.~\ref{fig:selection}. Entropy refers to the entropy-based selection used in prior TPT methods~\cite{shu2022test, Dynaprompt, C-TPT}. Cosine refers to the selection strategy~\cite{DiffTPT} which incorporates both prediction entropy and cosine similarity to the original image.
Entropy-based selection often focuses on background or misaligned regions, revealing its tendency to prioritize low-entropy predictions rather than semantically meaningful ones. Similarly, cosine-based selection often favors views that are visually similar to the original image but semantically misaligned, revealing its limitation in capturing meaningful object-level consistency.

Text anchor filtering in the proposed method alleviates this issue by emphasizing semantically faithful and class-discriminative regions aligned with textual descriptions.
Image anchor filtering complements the text-guided counterpart by selecting visually coherent and content-preserving views, guided by image anchors that capture the underlying distribution of test samples.
With the text and image filters , the method recovers cases where one modality alone fails.

\section{Conclusion}

We present a dual-modality anchor framework for test-time prompt tuning that grounds adaptation in both semantic and visual evidence.
Unlike prior entropy--based filtering approaches, our method constructs semantically enriched text anchors through alignment-weighted aggregation and complements them with incrementally updated image anchors derived from a prototype bank.
The dual anchors jointly provide alignment–confidence scoring that balances semantic fidelity and uncertainty, yielding more reliable view selection. Moreover, these anchors also serve as predictive sources whose ensemble provides a reliable target distribution, leading to stable adaptation.
Extensive experiments across multiple datasets demonstrate that our framework consistently outperforms existing test-time tuning methods, achieving robust performance.

\noindent \textbf{Acknowledgement.} This work was supported by the Institute of Information \& Communications Technology Planning \& Evaluation (IITP) grant funded by the Korea government (MSIT) [RS-2021-II211341, Artificial Intelligence Graduate School Program (Chung-Ang University)]
{
    \small
    \bibliographystyle{ieeenat_fullname}
    \bibliography{main}
}


\end{document}